\documentclass[letterpaper, 10 pt, conference]{ieeeconf}  

\IEEEoverridecommandlockouts                              

\usepackage{multirow}
\usepackage{wrapfig}
\usepackage{float}

\usepackage{graphicx}                               
\usepackage{amsmath}                                

\let\labelindent\relax          

\usepackage[inline]{enumitem}                       
\usepackage{textcomp}                               
\usepackage{gensymb}                                
\usepackage{xcolor}                                 
\usepackage[caption=false,font=footnotesize]{subfig}    
\usepackage{siunitx}                                
\usepackage{float}
\usepackage{array}
\usepackage{balance}
\usepackage{placeins}
\newcolumntype{M}[1]{>{\centering\arraybackslash}m{#1}} 

\usepackage{siunitx}

\usepackage{cite}

\title{\LARGE \bf
Using Neural Networks to Model Hysteretic Kinematics in Tendon-Actuated Continuum Robots
}

\author{Yuan Wang$^{1,\dagger}$, Max McCandless$^{1,\dagger}$, Abdulhamit Donder$^{1}$, Giovanni Pittiglio$^{1}$, \\ Behnam Moradkhani$^{2}$, Yash Chitalia$^{2}$ and Pierre E. Dupont$^{1}$
\thanks{$^{1}$Department of Cardiovascular Surgery, Boston Children’s Hospital, Harvard Medical School, Boston, MA 02115, USA.}%
\thanks{$^{2}$Department of Mechanical Engineering, University of Louisville, Louisville, KY 40292, USA}%
\thanks{*This work was supported by the NIH under grant R01HL124020}
\thanks{$\dagger$ Indicates shared first authorship.}
}

\begin{document}

\maketitle
\pagestyle{plain}


\vspace*{-5mm}

\begin{abstract}

The ability to accurately model mechanical hysteretic behavior in tendon-actuated continuum robots using deep learning approaches is a growing area of interest. 
In this paper, we investigate the hysteretic response of two types of tendon-actuated continuum robots and, ultimately, compare three types of neural network modeling approaches with both forward and inverse kinematic mappings: feedforward neural network (FNN), FNN with a history input buffer, and long short-term memory (LSTM) network. We seek to determine which model best captures temporal dependent behavior. We find that, depending on the robot's design, choosing different kinematic inputs can alter whether hysteresis is exhibited by the system. Furthermore, we present the results of the model fittings, revealing that, in contrast to the standard FNN, both FNN with a history input buffer and the LSTM model exhibit the capacity to model historical dependence with comparable performance in capturing rate-dependent hysteresis.

\end{abstract}

\vspace{1mm}

\section{INTRODUCTION}
\vspace{-1mm}

\begin{figure}[t]
  \centering
  \includegraphics[scale=1]{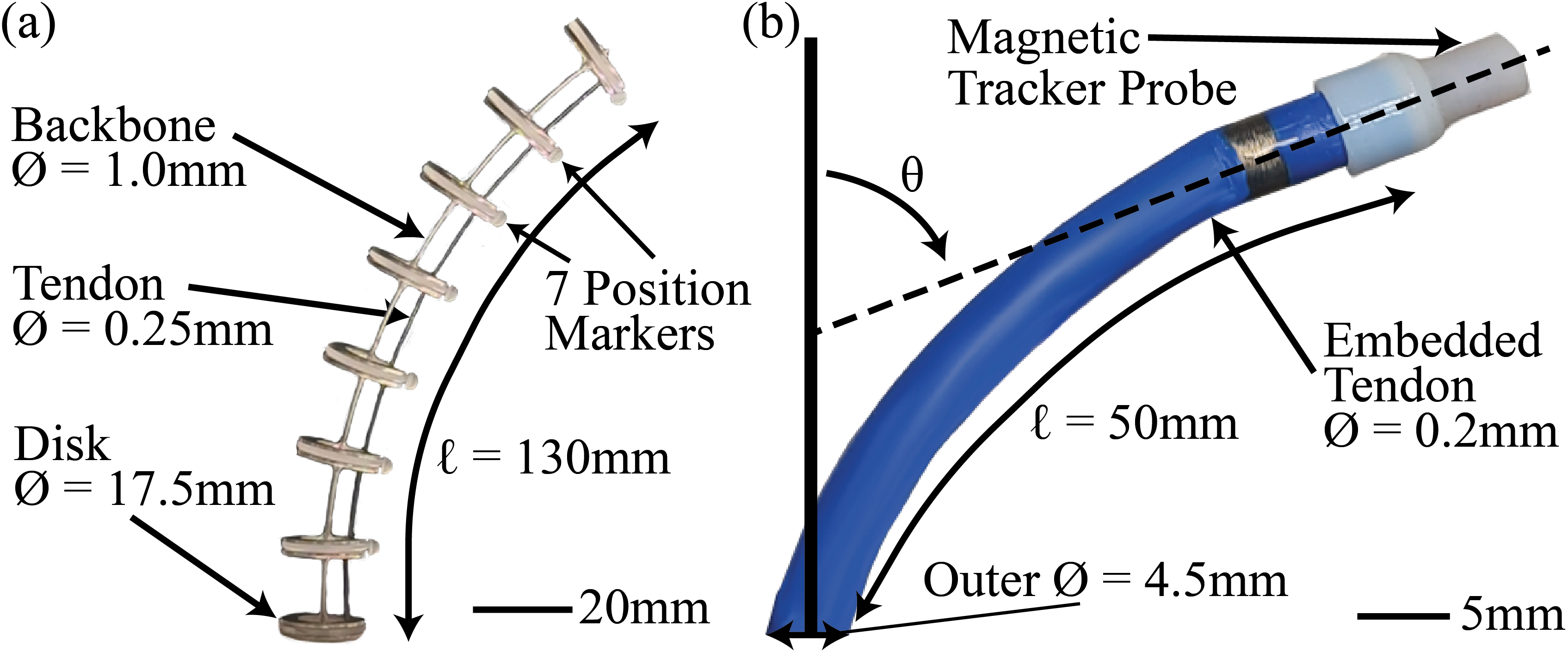}
  \vspace{-7mm}
    \caption{Tendon-actuated continuum robots. (a) Superelastic central backbone with spacer disks. (b) Clinical cardiac catheter.}
  \vspace{-5mm} 
  \label{fig:robots}
\end{figure}

\noindent
Since continuum robots produce a workspace through flexure of their components, modeling their kinematics is substantially more complex than for robots comprised of rigid links and discrete joints. Furthermore, since the flexure depends on the robot design, the modeling equations vary with robot type, e.g., concentric tube robots \cite{concentrictuberobots} versus tendon-actuated robots (Fig.~\ref{fig:robots})~\cite{tendon-actuated_robots}. Despite these differences, the general form of the equations is often similar: differential equations parameterized by arc length with boundary conditions split between the proximal and distal ends. 

Since these models are derived from mechanics, they are intuitive to the engineering mind and, when fitting a model to a specific data set, it is easy to interpret the meaning of the optimized parameters. Mechanics-based models do, however have several limitations. For example, they typically assume linear elasticity and ignore mechanical hysteresis and friction since these effects can be too complex to formulate and solve in real time~\cite{JHa1,JHa2}. 

These effects can be important. For example, it was shown in \cite{HSMR17} that the tip position of a 3-tube concentric tube robot can vary by 7\% of the robot length depending on the path in joint space by which the configuration is approached. Furthermore, as continuum designs become more complex, e.g., eccentric tube robots \cite{Eccentric_tube-robots-1,Eccentric_tube-robots-2} or telescoping tendon-actuated tubes \cite{telescopingtendonactuatedtubes}, the equations can become hard to solve in real time even when these effects are neglected. 

A number of approaches have been proposed to speed solution time and to include nonlinear effects (e.g., functional approximations \cite{concentrictuberobots} and nonparametric models \cite{Nonparametric_models}). In contrast, the modeling of hysteretic effects has received much less attention. While hysteresis models such as the Preisach and Bouc-Wen models \cite{PreisachModelandBOuc-Wenmodel} have been developed explicitly to reproduce hysteretic effects, it remains challenging to estimate model parameters based on data sets~\cite{Challengingtoestimatehystereticmodelparameters:}. 

With the explosion of interest in deep learning, neural networks are being applied as an alternative technique to mechanics-based modeling of continuum robot kinematics \cite{Paper3,article4}. As black box modeling schemes, they lack the intuitiveness of mechanics-based models, but they do offer the potential to reproduce neglected nonlinear and history-dependent (hysteretic) effects.

Hysteresis modeling can be considered as a form of sequence modeling, which is used to understand systems that exhibit dependencies on historical inputs. Sequence modeling is involved in many areas ranging from natural language processing (NLP), financial forecasting to DNA sequence Analysis \cite{RN103}. Feedforward Neural Networks (FNNs) and Recurrent Neural Networks (RNNs) are two fundamental forms of artificial neural networks that have been used for sequence modeling \cite{lipton2015critical} that can be applied to kinematic modeling.

For example, FNNs have been used to model the forward and inverse kinematics (tip position and orientation) of concentric tube robots~\cite{8594451,9981719}. For concentric tube robot shape estimation, the output of an FNN was used to compute the coefficients for a basis set of shape functions~\cite{Paper3}. In these examples, the FNNs are trained on kinematic data to produce a nonlinear input-output map and so can reproduce the nonlinear elastic behavior of the robot tubes. 

FNNs do not include any internal states or memory, however, and so they cannot be used to directly model hysteresis. One approach to get around this limitation is to redefine the inputs of the FNN to consist of the current input augmented by the prior $n-1$ input values where $n$ is selected to be sufficiently large to capture the hysteretic effects of interest. In this way, the FNN is trained to map a historical sequence of input values to predict a current output value. This approach has been used, for example, to model the inverse kinematics of a tendon-actuated robot~\cite{article4}.

In contrast to FNNs, RNNs are designed to model systems that exhibit dependencies on historical inputs. Consequently, they can directly be used to model both nonlinear and hysteretic effects that are neglected in mechanics-based models. A popular type of RNN is the Long Short-Term Memory (LSTM) model which overcomes the vanishing gradient problem to capture long-term dependencies more effectively \cite{10.1162/neco.1997.9.8.1735}.

For example, LSTMs have been used to model nonlinear elasticity and friction in the tendons of robots with rigid links \cite{CHOI2020102398} and in flexible endoscopic robots to estimate tendon force at the robot tip \cite{LI2019323}. They have also been used for modeling the combined input-output map of a pneumatic artificial muscle driving a tendon-actuated continuum robot~\cite{9360444}. Furthermore, LSTMs have been combined with Preisach model to learn the hysteresis characteristics of a single curved tendon-sheath mechanism \cite{KimRNNPreisach}.

The goal of this paper is to begin a systematic exploration of the application of neural networks to modeling tendon-actuated continuum robots. In Section II, the use of feedforward and recurrent neural networks is discussed with a focus on developing models that incorporate hysteretic behavior. Section III presents experimental modeling results for two tendon-actuated robot designs. Conclusions are presented in Section IV.

\section{Hysteretic Kinematic Modeling}

\noindent
Hysteresis is usually classified as rate independent and rate dependent. Applied to kinematic modeling, the former implies that a robot’s current configuration depends on its path history. In contrast, rate-dependent hysteresis implies that a robot’s configuration depends on its prior trajectory – its path as a function of time. 

\begin{figure}[t]
  \centering
  \includegraphics[scale=1]{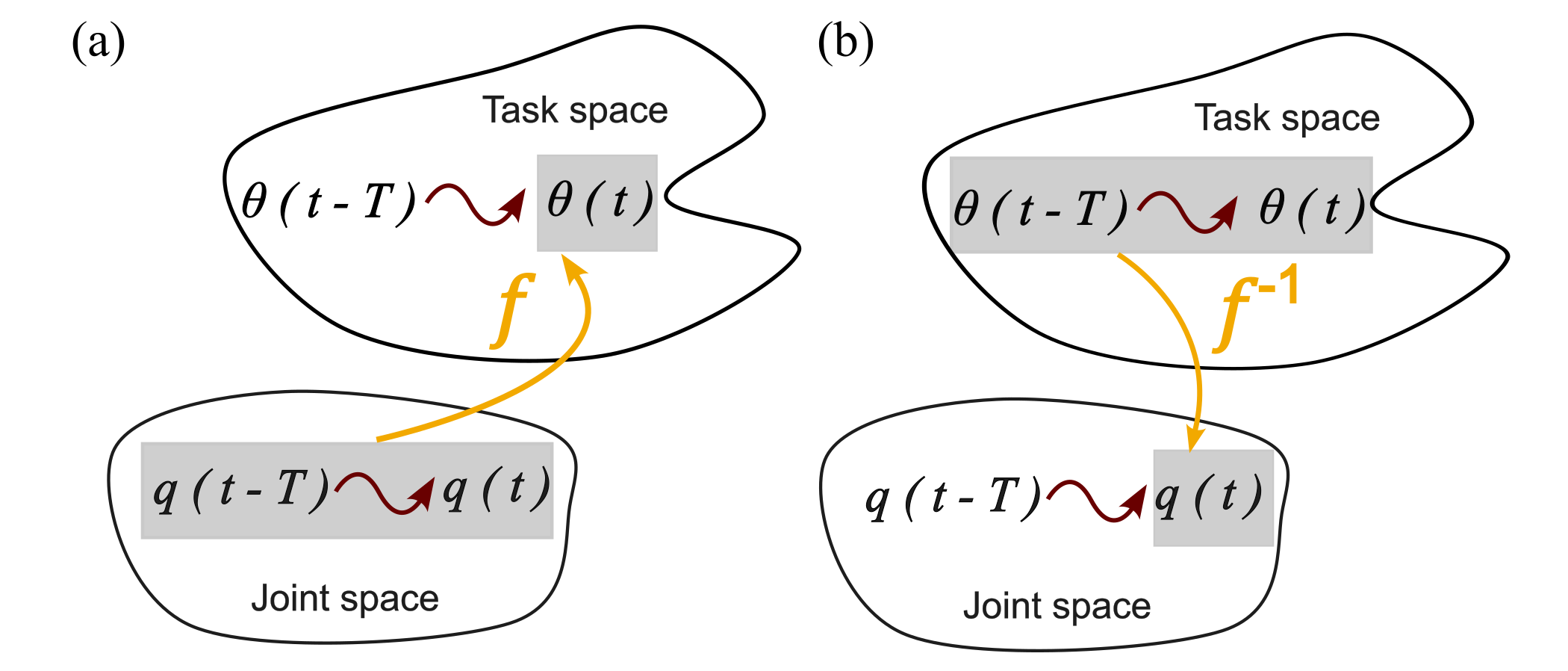}
  \vspace{-6mm}
    \caption{History-dependent kinematic mappings. (a) Forward kinematic map. (b) Inverse kinematic map.}
  \vspace{-4mm} 
  \label{fig:mappings}
\end{figure}

In either case, hysteresis requires that the forward and inverse kinematic solutions be computed by specifying not only the desired configuration, but also the path or trajectory of approach. If we assume that the motion can be parameterized by $t$, where $t$ could be time or some other representation of path length then the forward and inverse mappings are as shown in Fig.~\ref{fig:mappings} and can be written as:

\vspace{-2mm}

\begin{equation}
    \theta(t) = f(q(\tau)), \quad \tau \in [t - T, t]
    \label{eq:forward_kinematics}
\end{equation}
\vspace{-4mm}
\begin{equation}
    q(t) = f^{-1}(\theta(\tau)), \quad \tau \in [t - T, t]
    \label{eq:inverse_kinematics}
\end{equation}
\vspace{-4mm}

\noindent
here, $\theta$ represents robot tip position and orientation while $q$ is the vector of joint variables such as tendon displacements or tensions. The variable $T$ is defined as the maximum history length needed to predict the current configuration. 

While this formulation is useful for motion planning, kinematic computations for real-time control are typically performed in discrete time and are based on solving for incremental displacements. In this context, the equations above become \\


\vspace{-9mm}

\begin{equation}
    \theta[n] = f(q(m\cdot \Delta t)), \quad m \in [n-l+1, n-l+2, ..., n]
    \label{eq:discrete_forward_kinematics}
\end{equation}
\begin{equation}
    q[n] = f^{-1}(\theta(m\cdot \Delta t)), \quad m \in [n-l+1, n-l+2, ..., n]
    \label{eq:discrete_inverse_kinematics}
\end{equation}

\noindent
Here, $n$ is the discrete sample index and $\Delta t$ represents the sampling interval of sequential measurements. The index $m$ ranges from $n-l+1$ to $n$, where $l$ denotes the length of history samples used to predict the current configuration.

\subsection{Training Data}\label{trainingData}

\noindent
In order to train a model to reproduce hysteresis, kinematic data must be collected by sampling input-output paths or trajectories in configuration and task space. This differs from the usual approach used for physics-based models that involve collecting input-output data at a set of discrete points in configuration and task space. 

In addition, the input-output paths collected for training should be sufficiently exciting and of long enough duration to fully capture hysteretic behavior. In particular, hysteresis exhibits loops of varying size and so the training data should include loops covering the size range that is important for the intended clinical task. For single input – single output systems, one good approach proposed in \cite{9360444} is to use decaying sinusoids with different mean offsets as given in \eqref{eq:training}. The decaying sinusoids can capture frequency response and characterize hysteresis across diverse motion patterns.


\vspace{-3mm}

\begin{equation}
    \begin{split}
        q(t) &= q_{\text{max}}e^{-\tau t} \left(\sin\left(2\pi f_h t-\frac{\pi}{2}\right)+c \right) \\
        &\quad + q_{\text{offset}}, \quad 0\le t \le t_{\text{max}}
    \end{split}
    \label{eq:training}
\end{equation}

\noindent
For a single tendon system, $q$ can correspond to tendon displacement or tension. The measured output can be robot bending angle or tip displacement from the neutral axis. The parameters $q_{\text{max}}$, $c$ and $q_{\text{offset}}$ determine the amplitude, shape and offset of the signal, respectively. For a given frequency, $f_h$, the decay rate, $\tau$, and experiment duration, $t_\text{max}$, must be selected so that the experiment produces hysteresis loops over the magnitude range of interest. To generate sufficient training data, it is desirable to collect data at multiple frequencies. For model validation, data can be collected using (\ref{eq:training}) at additional frequencies.

\begin{figure*}[t]
  \centering
  \includegraphics[scale=1]{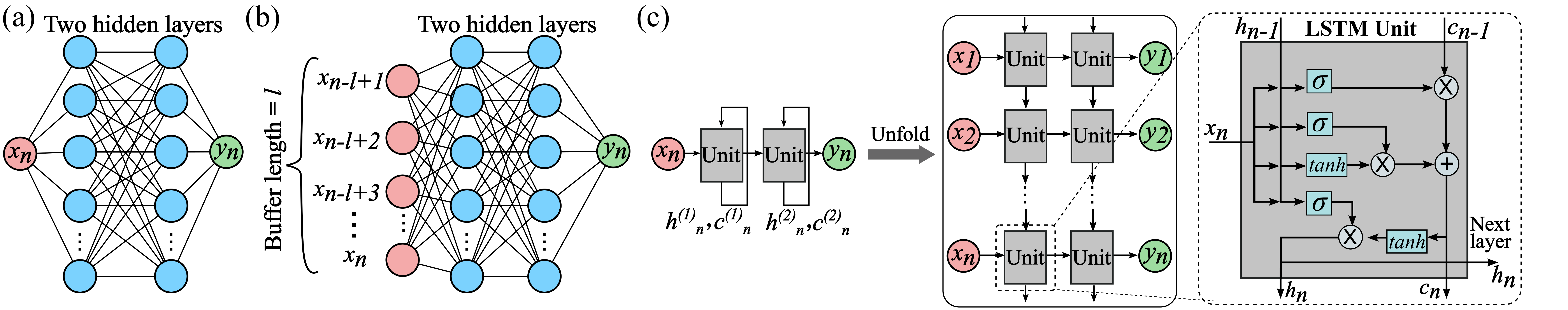}
  \vspace{-6mm}
    \caption{Neural network model structures. (a)~ FNN. (b)~ FNN with a history input buffer. (c)~ LSTM.}
  \vspace{-5mm} 
  \label{Learning Diagrams}
\end{figure*}

\subsection{Neural Network Models}

\noindent
In this paper, we compare three types of neural networks for modeling single tendon continuum robots: FNNs, FNNs with a history input buffer, and LSTMs.


\subsubsection{FNN} 
Standard FNNs use a unidirectional flow structure to learn paired input-output mappings. As shown in Fig.~\ref{Learning Diagrams}(a), FNNs used in this paper are structured with an input layer, followed by two hidden layers and an output layer. The number of nodes (neurons) in each layer is given as $[1, 64, 64, 1]$. The Rectified Linear Unit (ReLU) is used as the activation function to introduce nonlinearities. This FNN model can be expressed as follows:
\begin{equation}
    \begin{aligned}
    a^{(1)} &= \text{ReLU}(W^{(1)} \cdot x_{n} + b^{(1)}) & \\
    a^{(2)} &= \text{ReLU}(W^{(2)} \cdot a^{(1)} + b^{(2)}) & \\
    y_{n} &= W^{(3)} \cdot a^{(2)} + b^{(3)} 
    \end{aligned}
    \label{eq:FNN}
\end{equation}
where $W^{(i)}, i \in [1,2,3]$ are weight matrices, $b^{(i)}, i \in [1,2,3]$ are bias vectors, and $\text{ReLU}(x) = \max(0, x)$.

\subsubsection{FNN with a history input buffer} 
Unlike standard FNNs, where the current output is solely determined by the present input, FNNs augmented with a history input buffer (FNN-HIB) consider past input information, enabling it to capture temporal dependencies. The model input $x_{n}$ in \eqref{eq:FNN} is extended  to a vector \textbf{X} that encompass a set of samples from previous steps (Fig.~\ref{Learning Diagrams}(b)). This modified FNN is thus formulated as in \eqref{eq:FNN} with the following alteration:

\vspace{-2mm}

\begin{equation} 
    \begin{aligned} 
    a^{(1)} &= \text{ReLU}(W^{(1)} \cdot \mathbf{X} + b^{(1)}) 
    \end{aligned} 
    \label{eq:FNN_history} 
\end{equation}

\vspace{-1mm}

\noindent
and the input vector $\textbf{X} = [x_{n-l+1}, x_{n-l+2}, …, x_{n}]$.

\subsubsection{LSTM} 
LSTMs use a recurrent structure with memory units, as opposed to FNNs with fixed length of history input, to manage historical data over a long period of time. We apply a two-layer LSTM (Fig.~\ref{Learning Diagrams}(c)) for modeling in this paper. The global network structure can be formulated as \eqref{eq:Two-layer_LSTM}, with the LSTM units denoted as $\textit{lstm}$. Each input $x_n$ passes through two LSTM layers, followed by a linear layer to get the output at current step.

\vspace{-2mm}

\begin{equation} 
    \begin{aligned} 
    h^{(1)}_n &= \textit{lstm}(x_n) \\
    h^{(2)}_n &= \textit{lstm}(h^{(1)}_n) \\
    y_{n} &= W_{y} \cdot h^{(2)}_n + b_{y}
    \end{aligned}
    \label{eq:Two-layer_LSTM} 
\end{equation}

\vspace{-1mm}

\noindent
Here, the hidden state $h_n$ at each step serves as a mechanism for retaining memory. $W_{y}$ and $b_{y}$ are weight matrices and bias vectors of the last linear layer.





\subsubsection{Forward and inverse kinematic modeling}
Forward kinematics involves determining the position and orientation of the robot tip in task space based on the configuration in joint space. The model input $x_n$ in this context represents joint space configuration $q$, while the output $y_n$ corresponds to the resultant position or orientation $\theta$. 
Conversely, inverse kinematics involves determining the joint configurations necessary to achieve a desired position and orientation in task space. Here, the model's input $x_n$ represents the desired orientation $\theta$, and the output $y_n$ is the corresponding joint configuration $q$.








\section{EXPERIMENTAL MODELING}



\noindent
Experiments were conducted on the two single-tendon continuum robots of Fig.~\ref{fig:robots}. The robot of Fig.~\ref{fig:robots}(a) is comprised of a nickel titanium (NiTi) backbone with magnetic spacer disks inspired by the design of~\cite{FINAL_REFERENCE}. A NiTi tendon attached to the distal disk runs through holes in the spacer disks to the robot base. This robot design typifies what can be easily constructed in an academic research laboratory. The second robot of Fig.~\ref{fig:robots} is comprised of a polymer tube with embedded laser cut tubing and a stainless steel tendon wire running through a channel near the surface of the outer polymer layer. This design is representative of continuum robots produced for clinical use. 

\begin{figure}[t]
  \centering
  \includegraphics[scale=1]{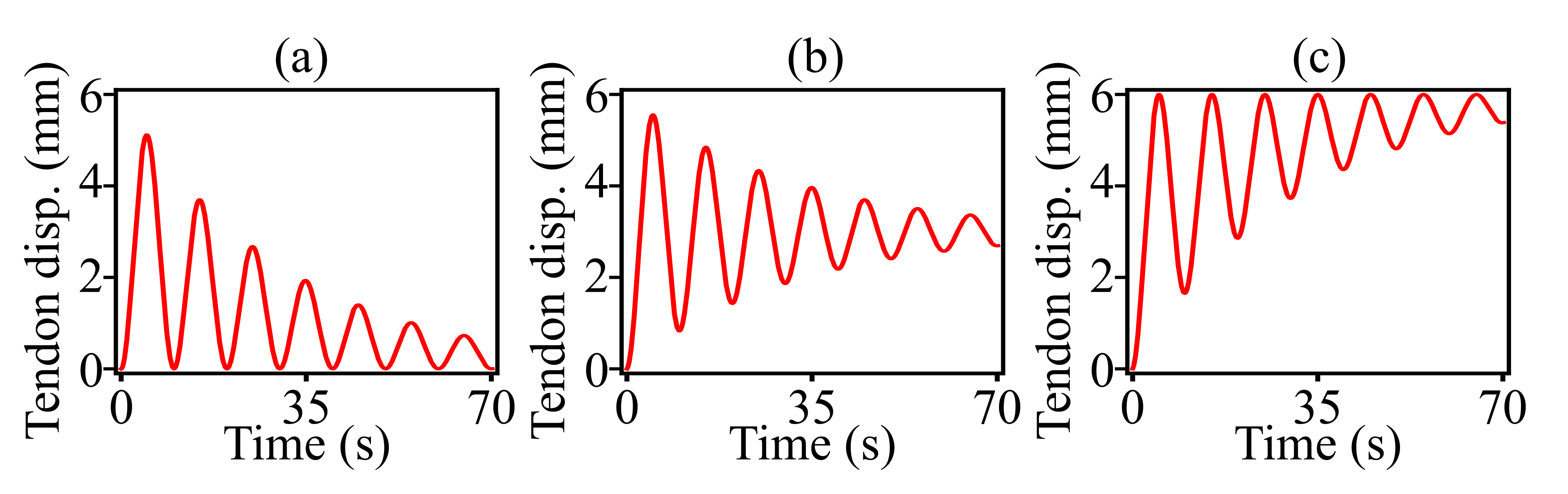}
  \vspace{-7mm}
    \caption{Training dataset tendon displacement sinusoidal function types conducted on the robot in Fig.\ref{fig:robots}(b) following (\ref{eq:training}) (each with $\tau$~=~$f_h \cdot \log(\frac{7}{6})$). (a)~0 baseline: $q_\text{max}$~=~\SI{3}{mm}, $c$~=~1, $q_\text{offset}$~=~0. (b)~Mid baseline: $q_\text{max}$~=~\SI{3}{mm}, $c$~=~0, $q_\text{offset}$~=~\SI{3}{mm}. (c)~End baseline: $q_\text{max}$~=~\SI{3}{mm}, $c$~=~-1, $q_\text{offset}$~=~\SI{6}{mm}.}
  \vspace{-0mm} 
  \label{Three type of sinusoids}
\end{figure}

Initial experiments were performed to determine if the robots exhibited hysteresis. If kinematic modeling required modeling hysteresis, a tendon pre-tensioning calibration procedure was developed to ensure that robot deflection was repeatable. Training and evaluation data sets were then collected using three forms of (\ref{eq:training}) as described in Fig.~\ref{Three type of sinusoids} (0 baselines (BL) Fig.~\ref{Three type of sinusoids}(a), Mid BL Fig.~\ref{Three type of sinusoids}(b), and End BL Fig.~\ref{Three type of sinusoids}(c)). This data was then used to compare the three model types: FNNs, FNN-HIB and LSTMs. The results for each robot appear in the subsections below. 

\subsection{Robot with Superelastic Central Backbone and Spacer Disks}\label{NiTiRobot}

\begin{figure}[t]
  \centering
  \includegraphics[scale=1]{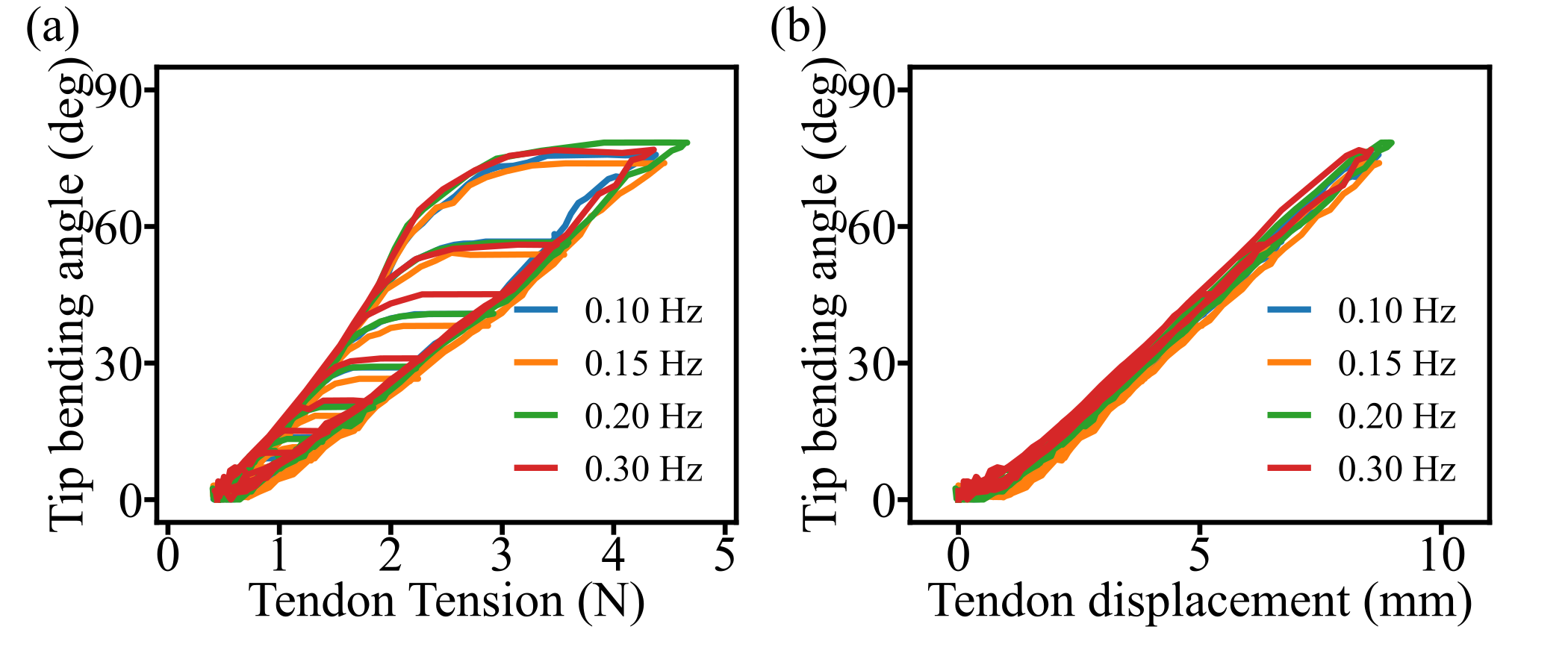}
  \vspace{-8mm}
    \caption{Alternative forward kinematic maps of robot shown in Fig.~\ref{fig:robots}(a). (a) Tip tangent angle as a function of tendon tension. (b) Tip tangent angle as a function of tendon displacement.}
  \vspace{-4mm} 
  \label{fig:yash}
\end{figure}

\noindent
The platform controlling the robot of Fig.~\ref{fig:robots}(a) consists of a Maxon DC motor (Maxon International Ltd. Switzerland) equipped with a reduction gearhead (gear ratio of 16:1) and a 100 counts per turn high precision encoder. For tracking the continuum robot position, four optical Vicon Vero cameras (Vicon Motion Systems Ltd. UK) are used to capture and track the optical markers (3~mm hemisphere facial markers) attached to multiple disks of the robot to establish the grouth truth position. 
Data was collected for the robot by applying tendon displacements following (\ref{eq:training}) with $q_\text{max}$~=~\SI{10}{mm}, $q_\text{offset}$~=~0, $c$~=~1 and frequencies, $f_h$, corresponding to 0.1-0.3~Hz. Since the robot controller can generate commands as either tendon displacements or tendon tensions, motor current was also logged during these experiments. 
The forward kinematic model can be considered either as mapping tendon displacement to tip tangent angle or tendon tension to tip tangent angle. The data for both maps is plotted in Fig.~\ref{fig:yash}. 

In physics-based models, tendon tension is typically viewed as the input~\cite{tendon-actuated_robots}. As shown in Fig.~\ref{fig:yash}(a), the forward kinematic map from tendon tension to bending angle exhibits rate-independent hysteresis. Based on this figure, it would appear that accurate kinematic control of the robot would require the use of a hysteretic model. As shown in Fig.~\ref{fig:yash}(b), however, the alternative kinematic model mapping tendon displacement to tip tangent angle does not exhibit hysteresis. In fact, the kinematic map from tendon displacement to tip angle can be represented as a linear equation. 




\subsection{Clinical Cardiac Catheter Robot}

\noindent
The platform controlling the robot of Fig.~\ref{fig:robots}(b) consists of a servo drive (Accelnet ACJ-055-09) controlling motors (Faulhaber Series 2342012CR) connected to the tendon pulleys with a gear ratio of 3.71 and encoder line drivers (Avago Technologies HEDL-5540 A06) used to track the tendon displacement. 
Prior to collecting training data, a set of pre-tensioning experiments were conducted in order to understand the tradeoff between the average deadband angle ($\phi$) induced by the pre-tension and the repeatability of the data collected between trials. 
Additionally, rate-dependance experiments were conducted to determine the frequency in which the hysteretic behavior of the system is variable (larger error in repeatability when compared to other frequencies). 
Following these system analyses, data was collected for the robot (using a magnetic tracking system as the ground truth (NDI 3D Guidance\textsuperscript{\textregistered{}} trakSTAR\textsuperscript{TM} and Model 180 sensor)) by applying tendon displacements following (\ref{eq:training}) with $q_\text{max}$~=~\SI{3}{mm}, three variations of $q_\text{offset}$ and $c$ (Fig.~\ref{Three type of sinusoids}(a-c)), and frequencies between 0.1-.05~Hz corresponding to tip displacement velocities of 6.5-32.5~mm/s. 
This training data was utilized to develop and compare the learning models at the frequencies tested as well as predicting the results of additional experiments run at intermediate frequencies of 0.15~Hz and 0.45~Hz.

\subsubsection{Tendon Pre-tension}

\begin{figure}[t]
  \centering
  \includegraphics[scale=1]{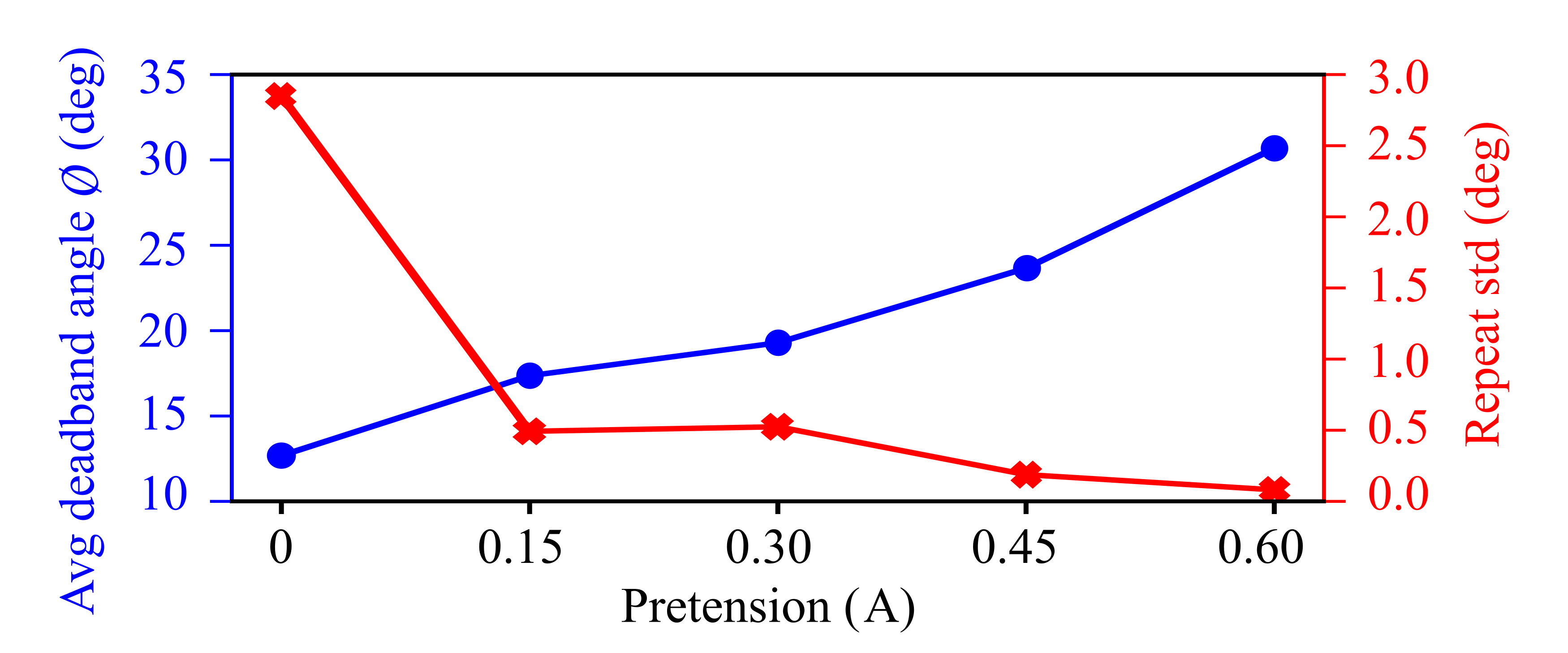}
  \vspace{-7mm}
    \caption{Effects of tendon pre-tension from 0-0.60~A (0-6.56~N). The plot shows that with increasing pre-tension there is an increase in repeatability (decrease in error between trials), but concurrently the deadband angle is increased resulting in less usable workspace.
    }
  \vspace{-5mm} 
  \label{Pretension Hansen Cath}
\end{figure}

\noindent
When the continuum robot is initialized, the slack in the tendon can vary greatly, which, for the same input tendon displacement, can result in over 15$^\circ$ of difference in tip angle. 
For this reason, a standard pre-tensioning approach was developed to remove the slack in the system.
For repeatability, the pre-tensioning was applied at a rate small enough to ensure quasistatic loading of the tendon up to a set motor current between 0-0.60~A (0-6.56~N) which was reached and maintained statically for $\approx$10~s before starting the experiment. 
To characterize the effects of pre-tension on the tendon-actuated catheter at varying amounts of initial tendon slack, experiments were conducted by applying tendon displacements following (\ref{eq:training}) with $q_\text{max}$~=~\SI{3}{mm}, $q_\text{offset}$~=~0 and $c$~=~1 (Fig.~\ref{Three type of sinusoids}(a)).
The results are shown in Fig.~\ref{Pretension Hansen Cath}, where the tradeoff between the increase in $\phi$ and the decrease in the error of repeated trials (average standard deviation of five trials) for increased pre-tension force is depicted. 
There is clearly a large benefit to the repeatability when pre-tensioning the system, which can be seen by the drop in error from 2.85$^\circ$ to 0.49$^\circ$ when moving from 0~A (0~N, no pre-tension) to 0.15~A (1.64~N). 
This relatively small increase to the pre-tension (0.15~A) also has minor effects on $\phi$ (only increased by 4.68$^\circ$), as compared to higher pre-tensioning values.
Therefore, the 0.15~A pre-tension would likely be considered sufficient in a clinical scenario where the maximum workspace size is preferred and the average error of 0.49$^\circ$ may be acceptable. For the purposes of this paper, however, we selected a pre-tension of 0.45~A (4.92~N) for the remainder of the experiments because we seek to understand and compare the capabilities of our learning models at predicting the tip angle of the system and, in turn, value the added repeatability of the tests at this larger pre-tension value (0.19$^\circ$). 
As compared to 0.60~A (6.56~N), 0.45~A represents the last data point where the tradeoff between increasing the repeatability and increasing $\phi$ is considered valuable for our modeling purposes (6.30$^\circ$ increase in $\phi$ from 0.15~A to 0.45~A).


\subsubsection{Hysteretic Rate Dependence} \label{ratedep}

\begin{figure}[t]
  \centering
  \includegraphics[scale=1]{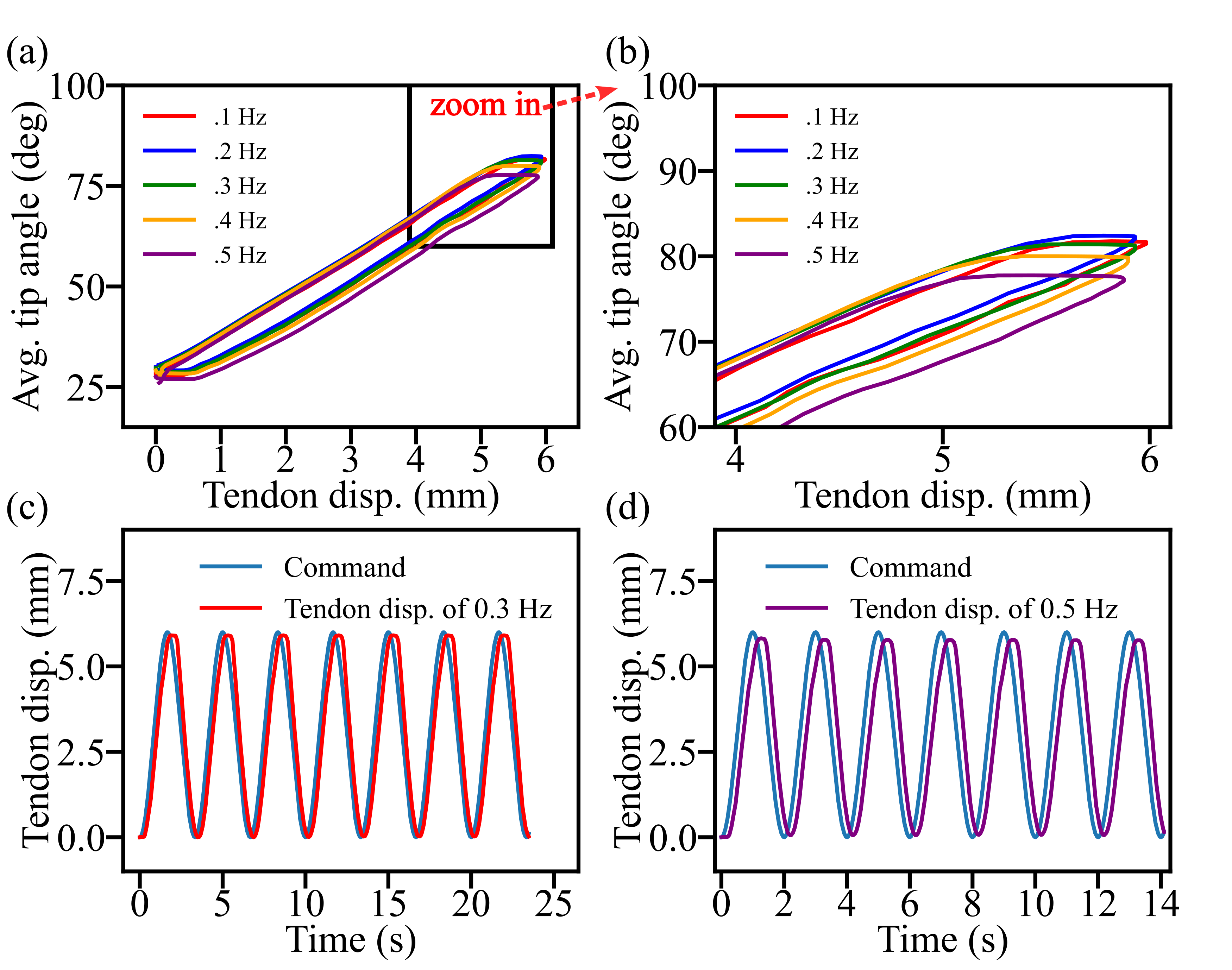}
  \vspace{-7mm}
    \caption{Hysteretic rate dependence. (a)~Tip angle displacement as a function of tendon displacement at five frequencies. (b)~Plot highlighitng the rate-dependent behavior of the system at frequencies higher than 0.3~Hz. (c)~Commanded versus actual tendon position at 0.3~Hz showing negligible difference between the two. (d)~Commanded versus actual tendon position at 0.5~Hz showing the actuator-induced lag of the system.    
    }
  \vspace{-4mm} 
  \label{Rate Dependant}
\end{figure}

The robot was tested at five different frequencies (0.1-0.5~Hz) to determine the frequency in which the system exhibits rate-dependent behavior. 
The tests consisted of seven non-decaying sinusoidal cycles with identical commanded tendon displacements ($q_\text{max}$~=~\SI{3}{mm}). 
The results can be seen in Fig.~\ref{Rate Dependant}(a) where the five trials of each frequency are averaged together and compared. 
If we focus on the behavior of the system at large tendon displacement (see subset of data in Fig.~\ref{Rate Dependant}(b)), we can see the rate-dependent nature of the robot is exhibited at a frequency greater than 0.3~Hz, which can be seen by the change in both the maximum bending angle achieved for the same input command and the width of the hysteresis loop at frequencies of 0.4~Hz and 0.5~Hz. 
When investigating this phenomenon further, it was determined that this rate-dependent hysteretic behavior (change in maximum bending angle achieved) likely arises more so from the motor dynamics (actuator-induced lag) than being a characteristic of the robot itself.
This actuator-induced lag can be seen in Fig.~\ref{Rate Dependant}(c-d), wherein Fig.~\ref{Rate Dependant}(c) shows the commanded tendon position and the actual position at 0.3~Hz with a negligible difference between the peaks of the two curves.
In contrast, the peak actual tendon position at 0.5~Hz (Fig.~\ref{Rate Dependant}(d)) lags behind the commanded tendon position and only reaches a maximum tendon displacement of $\approx$\SI{5.80}{mm}.
At a frequency of 0.5~Hz the tip of robot is moving at a rate of $\approx$~32.5~mm/s, which is much faster than clinically applicable for the system. 
It is likely the system, in practice, would be operating at frequencies well below the observed rate dependent behavior of 0.3~Hz (19.5~mm/s).
Regardless of this tip velocity consideration, Sect.~\ref{decaying:descriptions} will discuss the ability of the three neural network models tested to capture these rate-dependent behaviors.

\subsubsection{Comparison of Neural Network Models}\label{decaying:descriptions}



\paragraph{Data collection}

\begin{table*}
    \caption{Comparison of the performance of three neural network models with both forward and inverse kinematics.}
    \vspace{-2mm}
    \centering
    \resizebox{\textwidth}{!}{
    \begin{tabular}{|c||c|c|c|c|c|c||c|c|c|c|c|c|}
    \hline
    \multicolumn{1}{|c||}{{\textbf{\textbf{Forward Kinematic}}}} & \multicolumn{3}{|c|}{\textbf{0.15 Hz RMSE (deg)}} & \multicolumn{3}{|c||}{\textbf{0.45 Hz RMSE (deg)}} & \multicolumn{3}{|c|}{\textbf{0.15 Hz NRMSE}} & \multicolumn{3}{|c|}{\textbf{0.45 Hz NRMSE}} \\
    \cline{2-13}
    \multicolumn{1}{|c||}{\textbf{Models}} & \textbf{0 BL} & \textbf{Mid BL}& \textbf{End BL}& \textbf{0 BL} & \textbf{Mid BL}& \textbf{End BL} & \textbf{0 BL} & \textbf{Mid BL}& \textbf{End BL}& \textbf{0 BL} & \textbf{Mid BL}& \textbf{End BL}  \\
    \hline
    \textbf{FNN} &  2.824&  2.781&  2.858&  3.225&  3.061& 3.225& 5.2\%&4.7\%&4.4\%& 6.3\%& 5.3\%&5.0\%\\
    \cline{1-13}
    \textbf{FNN-HIB} &0.812&0.888&1.232&0.690&0.689&0.443& 1.5\%&1.5\%&1.9\%& 1.4\%& 1.2\%&0.7\%\\
    \cline{1-13}
    \textbf{LSTM} &1.016&1.140&1.463&0.713&0.641&0.415& 1.9\% &1.9\%&2.3\%& 1.4\%& 1.1\%&0.6\%\\ 
    \hline\hline
    \multicolumn{1}{|c||}{{\textbf{\textbf{Inverse Kinematic}}}} & \multicolumn{3}{|c|}{\textbf{0.15 Hz RMSE (mm)}} & \multicolumn{3}{|c||}{\textbf{0.45 Hz RMSE (mm)}} & \multicolumn{3}{|c|}{\textbf{0.15 Hz NRMSE}} & \multicolumn{3}{|c|}{\textbf{0.45 Hz NRMSE}} \\
    \cline{2-13}
    \multicolumn{1}{|c||}{\textbf{Models}} & \textbf{0 BL} & \textbf{Mid BL}& \textbf{End BL}& \textbf{0 BL} & \textbf{Mid BL}& \textbf{End BL} & \textbf{0 BL} & \textbf{Mid BL}& \textbf{End BL}& \textbf{0 BL} & \textbf{Mid BL}& \textbf{End BL} \\
    \hline
    \textbf{FNN} & 0.289&  0.263&  0.288&  0.304&  0.282& 0.310&5.7\%&4.7\%& 4.8\%& 6.2\%& 5.3\%&5.3\%\\
    \cline{1-13}
    \textbf{FNN-HIB} &0.099&0.090&0.105&0.090&0.082&0.099&1.9\% &1.6\%& 1.8\%& 1.8\%& 1.5\%&1.7\%\\
    \cline{1-13}
    \textbf{LSTM} &0.094&0.128&0.098&0.079&0.045&0.072&1.8\% &2.3\%& 1.7\%& 1.6\%& 0.8\%&1.2\%\\
    \hline
    \end{tabular}
    }
    \label{tab:model_comparison}
    \vspace{-4mm}
\end{table*}

As presented in Fig.~\ref{Three type of sinusoids}, decaying sinusoids with three different mean offsets were given as tendon displacements to gather the training datasets. Five frequencies [0.1, 0.2, 0.3, 0.4, 0.5] Hz of data for these three types of input were collected for training. 
Sect.~\ref{ratedep} showed that the catheter exhibits rate-dependent hysteresis when the sinusoidal frequency is greater than 0.3~Hz. Therefore, we adopted 0.15~Hz and 0.45~Hz sinusoidal waves to test the ability of modeling rate dependence. 
Sampling intervals of discrete data provides information on the time scale, which can be used for rate dependence modeling. All data was resampled at a fixed rate (25~Hz) to eliminate bias for model training and testing.

\paragraph{Modeling and training settings}
As demonstrated in Fig.~\ref{Learning Diagrams} and \eqref{eq:FNN}-\eqref{eq:Two-layer_LSTM}, the two types of FNNs share similar structure with 2 hidden layers, with each layer containing 64 nodes. The input layer dimension of the standard FNN is set to one, and in the FNN-HIB, the dimension is adjusted to match the chosen buffer size. The LSTM network adopts two recurrent layers, each with a hidden dimension set to 64.

The buffer size for the FNN-HIB was set 50 in this paper. Also, during the LSTM model training, the input tendon displacements are segmented into subsequences with a length of 50. To identify the start of sequence and make the initial 49 samples applicable, a list of flag values (-1) was prepended to the beginning of each sequence. Data were normalized between [0, 1] to facilitate training convergence.

To maintain consistency across all three models, identical training parameters were employed, including a learning rate of 0.001, a maximum of 500 training epochs, the Adam optimizer, Mean Square Error as the loss function, and a batch size of 16. These parameters are determined through empirical validation.

\paragraph{Model comparison}
The modeling performance was assessed using the Root Mean Squared Error (RMSE) and Normalized Root Mean Square Error (NRMSE), commonly used metrics in regression tasks. RMSE and NRMSE are calculated as:

\vspace{-2mm}


\begin{equation} 
    \begin{aligned} 
    \text{NRMSE} &= \frac{\text{RMSE}}{\text{max}(y) - \text{min}(y)}  &= \frac{\sqrt{\frac{1}{n} \sum_{i=1}^{n} (y_i - \hat{y}_i)^2}}{\text{max}(y) - \text{min}(y)}  
    \end{aligned} 
    \label{eq:RMSE} 
\end{equation}
\vspace{-2mm}

\noindent
RMSE provides an average measure of the residuals, effectively quantifying the deviation between predicted $\hat{y}_i$ and actual $y_i$ values. NRMSE further normalizes the RMSE, providing a relative measure of the prediction error that is independent of data scale.

\begin{figure}
  \centering
  \includegraphics[scale=1]{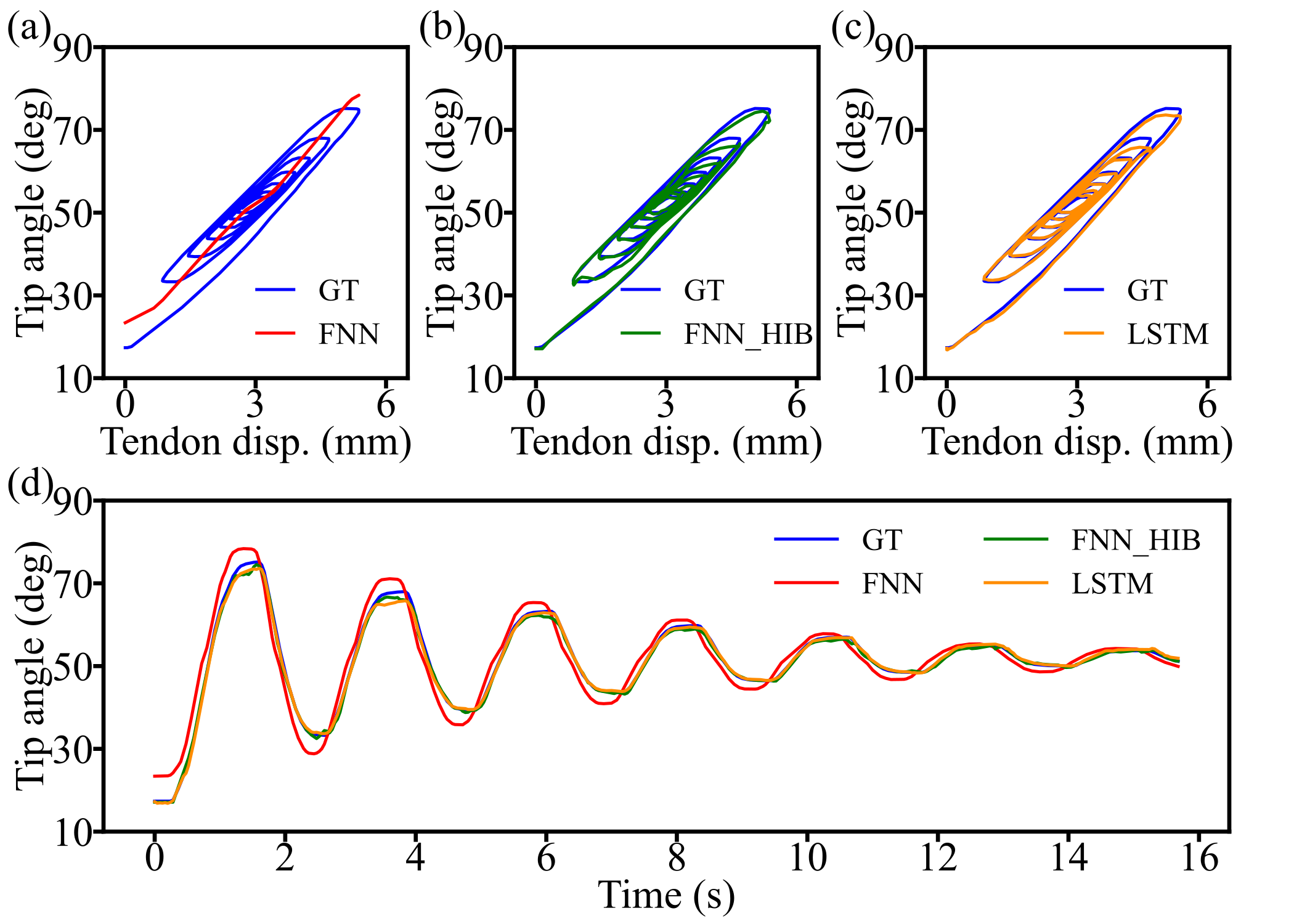}
  \vspace{-7mm}
    \caption{Forward kinematic testing results of three neural network models with Mid BL at 0.45~Hz. (a)~FNN (RMSE~=~3.061$^\circ$). (b)~FNN-HIB (RMSE~=~0.689$^\circ$). (c)~LSTM (RMSE~=~0.641$^\circ$). (d)~Ground truth tip angle (GT) versus model fittings over time.}
  \vspace{-5mm} 
  \label{Model_comparison}
\end{figure}

Sinusoidal tendon displacements with frequencies of 0.15 Hz and 0.45 Hz were applied to assess the performance of the models. Fig.~\ref{Model_comparison} shows one example of forward kinematics modeling using three types of neural network models. First, the model's predictions of tip bending angles are compared against the ground truth in relation to tendon displacements. As illustrated in Fig.~\ref{Model_comparison}(a), The FNN model fitted an approximately linear function between tendon displacement and tip bending angle (depicted by the red line), resulting in a significant error (RMSE: 3.061$^\circ$) in comparison to the ground truth, which is indicated by the blue line showing a hysteresis loop. In contrast, both FNN-HIB and LSTM models demonstrate the capacity to generate hysteresis loops, achieving accurate predictions with RMSE values of 0.689$^\circ$ and 0.649$^\circ$, respectively (Fig.\ref{Model_comparison}(b)-(c)). To make this comparison clearer, the predicted tip bending angles verses time were plotted in Fig.\ref{Model_comparison}(d). The FNN predicted motion (red line) consistently leads the real motion (blue line), revealing noticeable gaps at each peak. Contrastingly, both the FNN-HIB (green line) and LSTM (orange line) exhibit more accurate motion predictions.  

The quantitative evaluation results of all test dataset are listed in Table~\ref{tab:model_comparison}. The prediction errors of FNN-HIB and LSTM are generally close to each other, whereas the errors of FNN are approximately 2-5 times greater than those of the former two. The comparable modeling error for 0.15 Hz and 0.45 Hz motions indicates that FNN-HIB and LSTM models are able to model rate dependence effectively. It suggests that these neural network models can capture features in time scale by using data with fixed sampling rate. Also, the normalized error (NRMSE) of forward and inverse kinematic modeling for the same test motions (Table~\ref{tab:model_comparison}) show similar modeling results. This indicates that these three neural network models exhibit consistent performance in modeling the relationships between input parameters and desired outputs, showing their versatility and effectiveness in handling different tasks.

\vspace{-1mm}

\section{CONCLUSIONS}

\vspace{-1mm}

\noindent
This paper has explored the use of neural networks to model hysteresis in single-tendon continuum robots. An important observation is that the hysteretic behavior and thus the best way to model a robot is dependent on its design. For example, the robot of Fig.~\ref{fig:robots}(a) exhibits hysteresis if tendon tension is taken as the kinematic input, but not if the input is taken as tendon displacement. Furthermore, the hysteresis associated with tendon tension is rate independent. In this case, it is likely that the hysteresis with tendon tension is due to friction between the tendon and the disks and this friction did not have a velocity dependence in the measured frequency range. Since the tendon used in this robot was very stiff, it did not stretch in response to either the flexural forces or the friction forces and so did not produce a hysteretic kinematic model.

In contrast, the robot of Fig.~\ref{fig:robots}(b) exhibits hysteresis for both possible kinematic inputs, tendon tension and tendon displacement. And while the superelastic backbone of the robot of Fig.~\ref{fig:robots}(a) pulls it back to straight when the tendon is released, single-tendon clinical robot designs such as that of Fig.~\ref{fig:robots}(b) do not return to the straight configuration. Consequently, single tendon clinical designs are appropriate for tasks in which the robot workspace is located entirely off axis. For applications requiring the robot to be close to straight, antagonistic tendons are needed and, to date, neural network modeling of such designs has not been undertaken. 

The robot of Fig.~\ref{fig:robots}(b) was also observed to possess rate-dependent hysteresis above a specific frequency. It is likely that the frequency dependence arises from an actuator-induced lag and is not a characteristic of the robot. While neural network models can learn this lag, an alternative approach that is commonly used is to create a hybrid model structure combining neural networks with physics-based models. This approach is worthy of future consideration.

The modeling results demonstrate that FNNs without a history buffer cannot model historical dependencies, whereas FNNs with a history buffer and LSTM models demonstrate comparable ability in capturing such dependencies. LSTMs are somewhat simpler to implement since they require a single input at each time step while FNNs with a history buffer require implementation of the buffer of historical inputs. 







    
    
\bibliographystyle{IEEEtran.bst}
\typeout{}  
\bibliography{IEEEabrv.bib, references.bib}




\end{document}